\documentclass{bmvc2k}

\title{ReSynthDetect: A Fundus Anomaly Detection Network with Reconstruction and Synthetic Features}

\addauthor{Jingqi Niu}{stu_jqniu@sjtu.edu.cn}{1}
\addauthor{Qinji Yu}{cscyqj@sjtu.edu.cn}{1}
\addauthor{Shiwen Dong}{1481005290@sjtu.edu.cn}{1}
\addauthor{Zilong Wang}{zlwang@voxelcloud.net.cn}{2}
\addauthor{Kang Dang}{kdang@voxelcloud.net.cn}{2 \textdagger}
\addauthor{Xiaowei Ding}{dingxiaowei@sjtu.edu.cn}{1,2 \textdagger}

\addinstitution{
Department of Electronic Engineering\\
Shanghai Jiao Tong University\\
Shanghai, China
}
\addinstitution{
Voxelcloud Inc. \\
Chenhang Road, Minhang District,\\
Shanghai, China
}

\runninghead{Niu et al}{Reconstructive Synthetic Anomaly Detection}

\usepackage[T1]{fontenc}
\usepackage{graphicx} 
\usepackage{multirow} 
\usepackage{booktabs} 
\usepackage{amssymb}
\usepackage{mathrsfs}
\usepackage{amsmath} 
\usepackage{bm} 
\usepackage{graphicx}
\usepackage{makecell}
\begin{document}

\maketitle
\begin{abstract}
Detecting anomalies in fundus images through unsupervised methods is a challenging task due to the similarity between normal and abnormal tissues, as well as their indistinct boundaries. The current methods have limitations in accurately detecting subtle anomalies while avoiding false positives. To address these challenges, we propose the ReSynthDetect network which utilizes a reconstruction network for modeling normal images, and an anomaly generator that produces synthetic anomalies consistent with the appearance of fundus images. By combining the features of consistent anomaly generation and image reconstruction, our method is suited for detecting fundus abnormalities. The proposed approach has been extensively tested on benchmark datasets such as EyeQ and IDRiD, demonstrating state-of-the-art performance in both image-level and pixel-level anomaly detection. Our experiments indicate a substantial 9\% improvement in AUROC on EyeQ and a significant 17.1\% improvement in AUPR on IDRiD.
\end{abstract}

\vspace{-0.2cm}
\section{Introduction}
\label{sec:intro}
Deep Convolutional Neural Networks (CNNs) have made significant breakthroughs in various relevant fields of medical image analysis~\cite{huang2023revisiting, singh20203d, zhou2021review}. However, current fully supervised methods require a vast amount of annotated abnormal images. Obtaining such images can be challenging, particularly for rare diseases with low incidence rates. Conversely, collecting normal images is relatively easier. Therefore, recent research has focused on unsupervised anomaly detection methods~\cite{schlegl2019f, Niu2023, huang2022lesion2void} that identify anomalies in medical images through training only on normal images. Among medical imaging modalities, fundus image anomaly detection presents an especially challenging scenario. Retinal lesions come in various shapes, sizes, and textures. Learning them in an unsupervised manner can be difficult, particularly when they have indistinct boundaries and are visually similar to normal fundus tissues. Consequently, the development of an accurate and reliable method for detecting anomalies in fundus images is a critical area of ongoing research.

\begin{figure*}
  \centerline{\includegraphics[width=12cm]{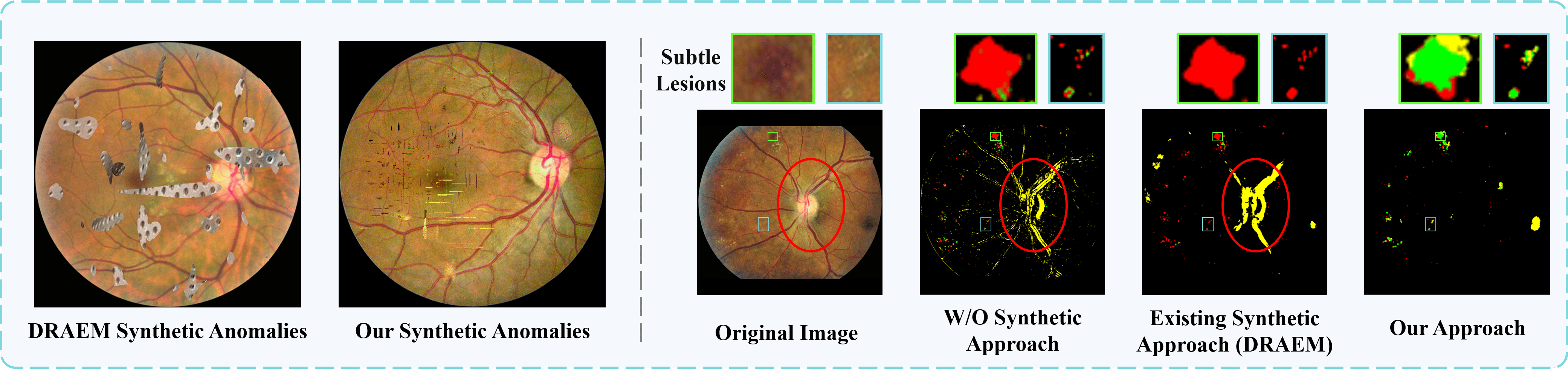}}
\caption{Our method generates more consistent lesions compared to existing methods (in left figure). Therefore, the existing methods have false positives in normal structures (see red circles), and our method has better detection performance (green area = true positives, yellow area = false positives, red area = missing). }
\vspace{-0.15cm}
\label{fig:intuation}
\end{figure*}

Currently, most anomaly detection methods rely on either reconstruction or representation based approaches~\cite{mao2020abnormality,ouardini2019towards,zhang2022multi}. However, these methods are typically trained solely on non-anomalous data and may not be optimized for discriminative anomaly detection. Consequently, they face challenges in learning abnormality representations and distinguishing  lesions that are not significantly different from normal tissues. This can lead to inaccurate detection of subtle lesions in fundus images and falsely identifying normal areas as anomalies, as shown in Fig.~\ref{fig:intuation}. To address this issue, recent studies~\cite{li2021cutpaste,zhao2021anomaly} have employed synthetic anomalies. Nevertheless, these techniques, such as DRAEM~\cite{zavrtanik2021draem}, may produce inconsistent anomalies that do not match the appearance of fundus images, which can be misleading. To overcome these limitations, we propose a novel approach that includes an anomaly generator capable of producing anomalies consistent with the appearance of fundus images. Additionally, to prevent overfitting to synthetic anomalies, we have implemented a reconstruction network which effectively reconstructs and models normal images. The reconstructive features produced by this network are combined with synthetic anomaly features to accurately localize any anomalies. As demonstrated in Fig.~\ref{fig:intuation}, our approach has been successful in detecting subtle fundus lesions while minimizing false positives by avoiding the misidentification of normal structures as anomalies.

Our proposed approach, named ReSynthDetect network which combines reconstruction and synthetic features, is designed for detecting fundus anomalies. The network is trained in two stages as shown in Fig.~\ref{fig:total}. In the first stage, a reconstruction network is trained on normal images, while in the second stage, an anomaly localization network is trained using artificially created anomalies. By incorporating information from the reconstruction network, the localization network accurately identifies anomalies while reducing reliance on synthetic anomalies. We create synthetic lesions by randomly selecting normal training images as source and target images. We augment the source images to create diverse lesions, which are then pasted onto random positions on the target images using self-mix~\cite{zhu2022selfmix}. This approach ensures consistency in the fundus images while simulating variations in real retinal lesions. Furthermore, we conducted comprehensive experiments on network architecture and found that, unlike previous literature~\cite{zavrtanik2021draem}, the best results for retina anomaly detection were obtained by combining the encoder features of the reconstruction network and the encoder features of the anomaly localization network. Through careful anomaly generation and network architecture selection, our proposed approach achieved state-of-the-art results on both the EyeQ \cite{fu2019evaluation} and IDRiD \cite{porwal2020idrid}  benchmark datasets.

\textbf{Contributions}. (1) We propose a new approach named ReSynthDetect network designed  for detecting anomalies in fundus images by combining reconstruction and synthetic features. (2) We introduce a novel anomaly generator that can produce diverse and consistent synthetic anomalies in fundus images. (3) Our proposed approach achieves state-of-the-art results on two benchmark retinal datasets, EyeQ and IDRiD, with a 9\% improvement in AUROC for image-level anomaly detection on EyeQ and a 17.1\% improvement in AUPR for pixel-level anomaly localization on IDRiD.

\section{Related Work}
Most of the existing anomaly detection methods are reconstruction-based \cite{gong2019memorizing,baur2021autoencoders,zhou2020encoding}, which train  models to reconstruct normal data during training and detect anomalies by calculating the reconstruction error. fAnoGAN in \cite{schlegl2019f} applies an adversarial network for normal image reconstruction and calculates anomaly scores by reconstruction error. WDMT in \cite{zhang2022multi}  use a weight-decay skip connection strategy for reconstruction network and integrating an auxiliary task of the histogram of oriented gradients prediction to improve feature representation learning. Lesion2Void in \cite{huang2022lesion2void} masks out normal patches and trains a reconstruction model based on the correlation with neighboring pixels to distinguish anomalies. However, these methods often  have relatively large reconstruction errors in normal retinal structures such as optic disc, cup, and blood vessels, resulting in potential false positives \cite{mao2020abnormality}. 

To solve this problem, some recent representation-based methods have been proposed \cite{roth2022towards,ouardini2019towards,defard2021padim} , which compute anomaly scores based
on the similarity of the features between the test and normal samples. ReSAD in \cite{Niu2023} extracts features by a pre-trained model  and proposes a spatial and region module for local and long-range anomaly detection. MKD in \cite{salehi2021multiresolution} applies knowledge distillation between a pre-trained source network and a smaller cloner network and calculates feature similarity as anomaly scores. Nevertheless, representation-based methods lack the reference of abnormal features, making it challenging to detect subtle lesions in fundus images.

A few works attempt to utilize synthetic anomalies for anomaly detection \cite{li2021cutpaste, schluter2022natural,zhao2021anomaly}.  DRAEM \cite{zavrtanik2021draem} trains a reconstructive network on synthetic anomalies and utilizes a discriminative network to detect deviations from synthetic and reconstructed images as anomalies. However, the reconstructed images may contain  deviations in normal structures, which can be falsely detected as anomalies. Our work is built on a similar architecture as described in~\cite{zavrtanik2021draem}, but we introduce two key technical differences to overcome its limitations. Firstly, we propose a novel anomaly generator that can produce  diverse and consistent synthetic anomalies in fundus images. Secondly, unlike previous literature~\cite{zavrtanik2021draem}, we achieve the best results for retina anomaly detection by combining the encoder features of both the reconstruction network and the anomaly localization network.


\vspace{-0.3cm}
\section{Method}
\begin{figure*}[t]

\centerline{\includegraphics[width=10.5cm]{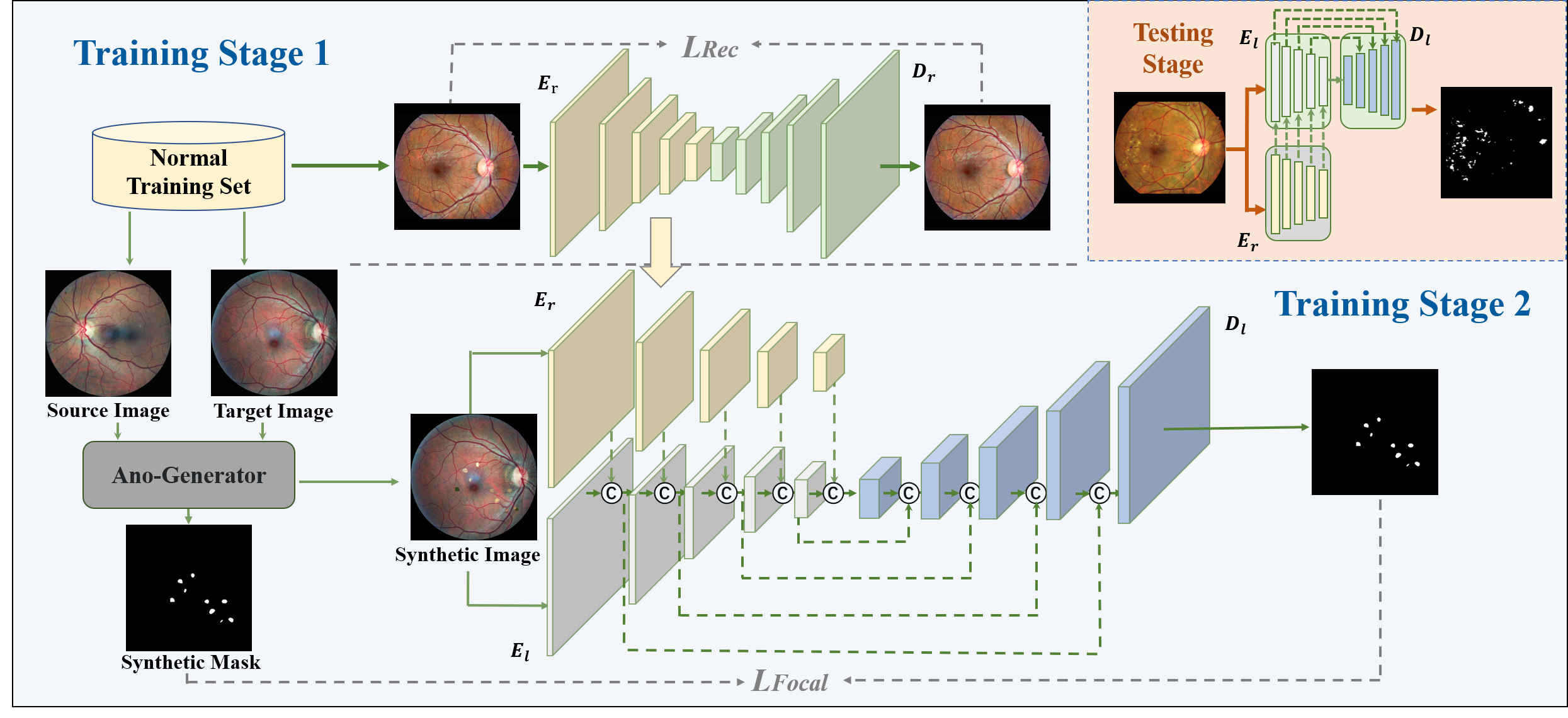}}
\caption{ \textbf{Overall Pipeline}: In the first training stage,  we train a reconstruction network to reconstruct normal images in order to obtain its encoder  $E_r$ as a feature extractor. In the second stage,  we concatenate the reconstructive feature extract from $E_r$ to a U-shape localization network  for synthetic anomalies localization. Finally, we test our model in the testing set by utilizing the localization network with reconstructive features extracted by $E_r$.} 

\label{fig:total}
\vspace{-0.4cm}
\end{figure*}

The pipeline of our proposed method is illustrated in Fig. \ref{fig:total}. In the first stage, an reconstruction network is trained on normal training images (Sec. \ref{Autoencoder}), and its encoder is utilized as a feature extractor in the subsequent training stage. In the second stage, a reconstructive feature-guided anomaly localization network is trained using synthetic anomalies (Sec. \ref{Localization}). The synthetic anomalies are obtained by a consistent anomaly generator (Sec. \ref{ano-generator}).

\vspace{-0.3cm}
\subsection{Reconstruction Based Feature Extractor}
\label{Autoencoder}



Relying solely on synthetic anomalies can result in overfitting to their specific patterns. Previous work, DRAEM \cite{zavrtanik2021draem}, combines reconstructed images with synthetic anomalies to identify deviations as anomalies. However, reconstructed images may contain deviations in normal structures, leading to false positives. To overcome this, we train a reconstruction network as a feature extractor, which mitigates overfitting to synthetic anomaly patterns and avoids false positives in normal structures.

In the first training stage, an autoencoder is utilized as the reconstruction network, with the objective of reconstructing the normal fundus images which serve as the input of the network. This process enables the extraction of reconstructive features, which are subsequently utilized in the second stage of training 

Formally, the reconstruction network comprised of an encoder $E_r$ and a decoder $D_r$, is trained on the normal training fundus image $\bm{I}$. We utilize the L2 loss as the reconstruction loss, which can be calculated as follow:


\begin{equation}
\label{equ:rec}
    L_{Rec} =  \Vert D_r(E_r(\bm{I})) -  \bm{I} \Vert^2_2.
\end{equation}

Once the training of the autoencoder is completed, its parameters are fixed and will no longer be changed in the following stage. The encoder $E_r$ of the reconstruction network will be used as the extractor of the reconstructive feature, while the decoder $D_r$ will be discarded.





\vspace{-0.3cm}
\subsection{Reconstructive Features Guided Localization Network}
\vspace{-0.15cm}
\label{Localization}
Due to the lack of real anomaly samples in the training phase, the network needs to be trained on proxy tasks. We utilize the localization of synthetic anomalies as the proxy task to train the network.

In the second training stage, a reconstructive feature-guided localization network is trained on the synthetic image $\bm{I_G}$ and its corresponding mask $\bm{M_G}$ (see Sec. \ref{ano-generator}).  We apply a U-shape model \cite{ronneberger2015u} with skip connections as the localization network, which consists of an encoder $E_l$ and a decoder $D_l$, to localize the synthetic anomalies. We  extract both reconstructive features extracted by $E_r$ and localization features extracted by $E_l$ in the synthetic image $\bm{I_G}$ and concatenate  them in each layer.

More specifically, at each layer $i$ of the encoders $E_r$ and $E_l$, we extract the corresponding reconstructive features $\bm{F_r^i}$ and localization features $\bm{F_l^i}$, respectively, and $\bm{F_c^i}$ is obtained by concatenating them  as $\bm{F_c^i} = concat (\bm{F_r^i}, \bm{F_l^i})$. Subsequently,  $\bm{F_c^i}$ is used as input in the subsequent layers of encoder $E_l$ and decoder $D_l$ for the localization task.

The Focal Loss \cite{Azhar2020} is introduced during training to alleviate the imbalance between the normal pixels and anomalous pixels. It is expressed as follows:
\begin{equation}
L_{Focal} = \left\{
\begin{aligned}
&-(1-p)^{\tau} \log(p), &M_G^{x,y} = 1, \\
&-p^{\tau} \log(1 - p) , &M_G^{x,y} = 0. 
\end{aligned}
\right.
\end{equation}
Here, $p$ represents the  probability of anomaly at position $(x,y)$ predicted by the model, and $\tau$ is a tunable focusing parameter, and set to $2$ in this paper.  

\vspace{-0.3cm}
\subsection{Consistent Anomaly Generator }
\vspace{-0.15cm}
\label{ano-generator}

\begin{figure*}[t]
\centerline{\includegraphics[width=10cm]{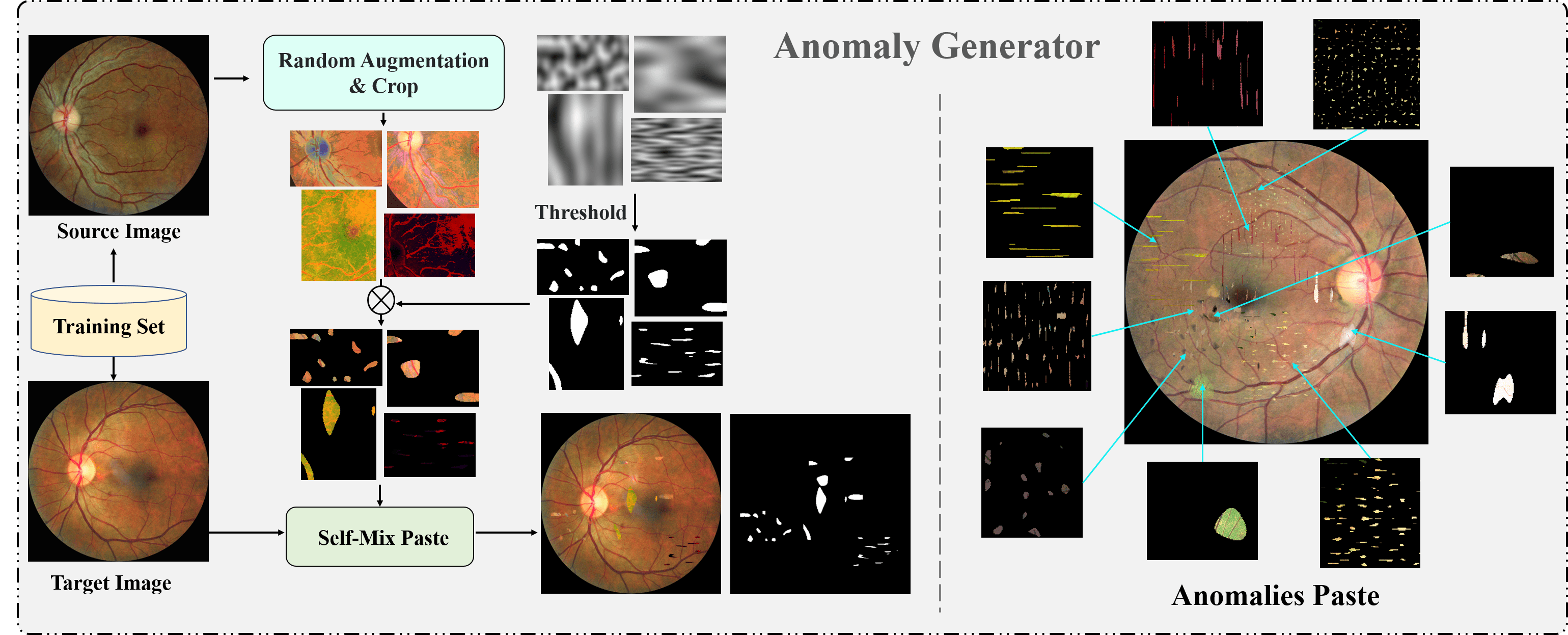}}
\caption{\textbf{Anomaly Generator}: 
The left image shows the process of the anomaly generator, which  creates lesions from a source image and pastes the generated lesions onto the target image. The image on the right demonstrates that the proposed method is capable of generating lesions with various shapes and textures. These lesions can be pasted onto random locations of the target images without disrupting the continuity of the image. More synthetic images can be found in the Supplementary.}
\vspace{-0.5cm}
\label{fig:SCAG}
\end{figure*}

As depicted in Fig. \ref{fig:SCAG}, our approach incorporates an anomaly generator that generates lesions based on the source image and subsequently pastes the generated lesions onto the target image. In order to maintain the consistent appearance between normal retinal images and synthetic anomalies, 
we randomly sample a source image $\bm{I_s}$ and a target image $\bm{I_t}$ from the normal training set. To generate a variety of texture anomalies, we randomly select three augmentation methods from our pool of candidates, which include sharpening, solarizing, gamma contrast enhancement, hue change, color temperature alteration, auto-contrast, and random color shifting. We apply these augmentations to the source image  $\bm{I_s}$ and produce an augmented image. Subsequently, the anomaly generator performs a random cut to obtain crop $\bm{C_s}$ of variable size from the augmented source image at a random location to generate lesions.  

To generate anomalies with diverse shapes, we use Perlin noise, a type of gradient noise commonly employed in computer graphics. Our anomaly generator utilizes a Perlin noise generator \cite{perlin1985image} to produce Perlin noise $\bm{P_n}$ of the same size as $\bm{C_s}$. Subsequently, a thresholding process is applied to generate a binary mask $\bm{P}$ from $\bm{P_n}$.


Directly pasting the augmented source crop $\bm{C_s}$ onto target images can potentially introduce inconsistencies in the boundaries of the pasted lesions. To address this issue, we use self-mix paste module, which uses Euclidean Distance Transform algorithm \cite{felzenszwalb2012distance} to compute the distance between each pixel in the  mask $\bm{P}$ and its nearest background pixel, generating a distance map $\bm{D}$.  Subsequently, fusion weights map \bm{$W$} are generated according to the  Eq. \ref{equ:concat}: 

\begin{equation}
\label{equ:concat}
    \bm{W} = (1- \alpha) \times \frac{\bm{D}-  min(\bm{D})}{max(\bm{D}) - min(\bm{D})} + \alpha .
\end{equation}
where $\alpha$ is the scaling factor, and set to $0.7$ in this paper. 

Next, self-mix paste module selects a crop \bm{$C_t$} in the target image with a random location and fuse the \bm{$C_s$} and \bm{$C_t$} with $\bm{W}$ and 
\bm{$P$}  according to Eq. \ref{equ:selfMix}.  Notably, \bm{$C_s$} and \bm{$C_t$} have the same size but may not be located at the same position.


\begin{equation}
\begin{gathered}
\label{equ:selfMix}
    \bm{C} = (\bm{P} \odot \bm{W}) \odot \bm{C_s} + (1- \bm{P} \odot \bm{W}) \odot \bm{C_t} ,\\
    \bm{C_m} = \bm{P},
\end{gathered}
\end{equation}
where $\odot$ denotes the element-wise product, $\bm{P} \odot \bm{W}$ denote a smoothing mask.


In  Eq. \ref{equ:selfMix}, $\bm{C}$ denotes the generated anomaly crop, the   $\bm{C_m}$ denotes the corresponding mask. Finally, the anomaly generator can obtain the synthetic image $\bm{I_G}$ and corresponding mask  $\bm{M_G}$ through $\bm{C}$ and $\bm{C_m}$, which is utilized in previous training.

\vspace{-0.2cm}
\section{Experiments}


\vspace{-0.1cm}
\subsection{Experimental Protocol}
\noindent\textbf{Datasets.} For evaluation, we used two public datasets: EyeQ \cite{fu2019evaluation} for image-level anomaly detection and IDRiD \cite{porwal2020idrid} for pixel-level anomaly localization. We applied Contrast Limited Adaptive Histogram Equalization (CLAHE) \cite{zuiderveld1994contrast} with a ClipLimit of 2 and a GridSize of 8 to enhance image contrast while preserving local details. The input image size for all datasets was standardized to 768 × 768 for consistency.
\begin{itemize}
\item{EyeQ}: The EyeQ \cite{fu2019evaluation} dataset is a subset of the EyePACS \cite{diabetic-retinopathy-detection}  dataset used for grading diabetic retinopathy (DR). EyeQ consists of 28,792 fundus photographs with DR grading annotations and corresponding image quality labels. The images in the EyeQ dataset are classified into "good", "usable", or "reject" categories based on the image quality, and only the images classified as "good" are used in our experiments. The DR disease severity in the EyeQ dataset is divided into five grades: 0 (normal), 1 (mild), 2 (moderate), 3 (severe), and 4 (proliferative) \cite{lin2020sustech}. Following \cite{huang2022lesion2void}, images with level 0 are considered as normal images, and images with grades 1 - 4 are considered as abnormal images. During the training phase, we  utilizes all the  normal images in  training set from the EyeQ dataset, which consist of 6,324 normal images. For testing, we used the complete testing set comprising 8,470 images from the EyeQ dataset. 

\item{IDRiD}: We used the Indian Diabetic Retinopathy Image Dataset (IDRiD) \cite{porwal2020idrid} dataset, which contains highly precise DR lesion  masks and is commonly used as a benchmark dataset for  lesion localization tasks. Specifically, we used 134 normal retinal images as the training set, 32 normal retinal images and 81 abnormal retinal images with finely annotated DR lesions as the testing set. The abnormal image contains  four types of retinal lesions with fine annotated masks, including microaneurysms (MA), soft exudates (SE), hard exudates (EX), and hemorrhages (HE). 
\end{itemize}

\vspace{-0.2cm}
 \noindent\textbf{Implementation Details.} The codes are implemented using PyTorch on a single NVIDIA RTX 3090 GPU with 24GB memory. The initial learning rate is set to 5e-5 with a cosine learning rate decay, reaching a minimum learning rate of 2.5e-5.  Additionally, a warm-up strategy with a duration of 50 epochs is implemented. For image-level anomaly detection, we compute the anomaly score by averaging the highest predicted anomaly probabilities of the top 10 pixels in the model's output.
 
 \noindent\textbf{ Evaluation Metric.} We evaluate the performance using the standard metric for anomaly detection, AUROC, for both image-level anomaly detection and pixel-level anomaly localization.  However, the AUROC can not precisely reflect the localization result especially in fundus images anomaly detection.  The reason is that  false positive rate is dominated by the \emph{a-priori} very high number of non-anomalous pixels and is thus kept low despite of false positive detections. We thus additionally report the pixel-wise Area Under the Precision-Recall curve (AUPR), which is more realistic  for the lesion localization performance \cite{kersting2013machine, saito2015precision}, especially for retinal lesion localization performance \cite{porwal2020idrid} because it is more appropriate for highly imbalanced classes. Besides, we also evaluate the performance of the method by balanced accuracy (ACC) on pixel-level anomaly detection in order to partially mitigate the issue of imbalanced distribution of positive and negative samples.

\begin{center}
\begin{table}[th]
\centering
\begin{tabular}{c|c|c|c|c|c }
\hline
Method & 0vs1 & 0vs2 & 0vs3 & 0vs4 & 0 vs all \\
\hline
fAnoGAN \cite{schlegl2019f} & 0.508 & 0.491 & 0.525 & 0.577 & 0.514 \\
MKD \cite{salehi2021multiresolution} & \textbf{0.582} & 0.547 & 0.623 & 0.706 & 0.546\\
DRAEM \cite{zavrtanik2021draem} & 0.585 & 0.658 & 0.742 & 0.716 &  0.614 \\
Lesion2Void \cite{huang2022lesion2void} & 0.567 & 0.625 & 0.877 &  0.902 & 0.632\\
Ours & 0.556 & \textbf{0.764} & \textbf{0.941}&  \textbf{0.919} & \textbf{0.722}  \\
\hline
\end{tabular}
\caption{Image Level Anomaly Detection Result for EyeQ in AUROC.}
\vspace{-0.2cm}
\label{tab:result_image}
\end{table}
\end{center}

\vspace{-1cm}
\begin{center}
\begin{table}[th]
\centering
\begin{tabular}{c|c|c|c}
\hline
Method   & AUROC &   ACC & AUPR  \\\Xhline{1 pt} 
fAnoGAN \cite{schlegl2019f}  & 0.756 & 0.686 & 0.048\\
MemAE \cite{gong2019memorizing}  & 0.749 & 0.596  & 0.058\\
WDNet \cite{zhang2022multi}  & 0.775 & 0.566 & 0.075 \\
DRAEM \cite{zavrtanik2021draem}  & 0.827 & 0.747 & 0.100 \\ 
ReSAD \cite{Niu2023}  & 0.905 & 0.819 & 0.256 \\

Ours  & \textbf{0.931} & \textbf{0.859} & \textbf{0.427}\\
\hline
\end{tabular}
\caption{Pixel Level Anomaly Detection Result for IDRiD.}
\vspace{-1cm}
\label{tab:result_pixel}
\end{table}
\end{center}

\vspace{-0.2cm}
\subsection{Comparisons with the State of the Arts}
\label{SOTA_Rse}
\textbf{Image-level Anomaly Detection.}  Following \cite{huang2022lesion2void}, we compare our proposed method with multiple SOTAs: two reconstruction-based
methods: fAnoGAN~\cite{schlegl2019f} and Lesion2Void \cite{huang2022lesion2void}, a synthetic anomalies based method DRAEM~\cite{zavrtanik2021draem}, a representation-based method MKD~\cite{salehi2021multiresolution} as the baseline model for image-level anomaly detection.

Tab. \ref{tab:result_image} quantitatively compares our model with baselines in the image-level anomaly detection on EyeQ. Grade 0 is considered as a normal image. For the comparisons of 0 vs 1, 0 vs 2, ..., 0 vs 4, we consider only DR graded images from grade 1 to grade 4 as abnormal images. For the comparisons of 0 vs all, we use all abnormal images with DR grades from 1 to 4 for anomaly detection. Our approach surpasses all the baselines in  0 vs all grade experiments and surpasses  the previous best SOTA method by 9\%, which demonstrates that our method can achieve state-of-the-art performance in image-level retinal anomaly detection.

 \noindent\textbf{Pixel-level Anomaly Localization.} We compare our method with multiple SOTA methods: three reconstruction-based methods: fAnoGAN \cite{schlegl2019f}, MemAE \cite{gong2019memorizing} and WDMT-Net \cite{zhang2022multi}, a synthetic anomalies based method DRAEM \cite{zavrtanik2021draem} and a representation-based approach  ReSAD \cite{Niu2023} as our baseline models.

\begin{figure*}[t]
  \centerline{\includegraphics[width=12.5cm]{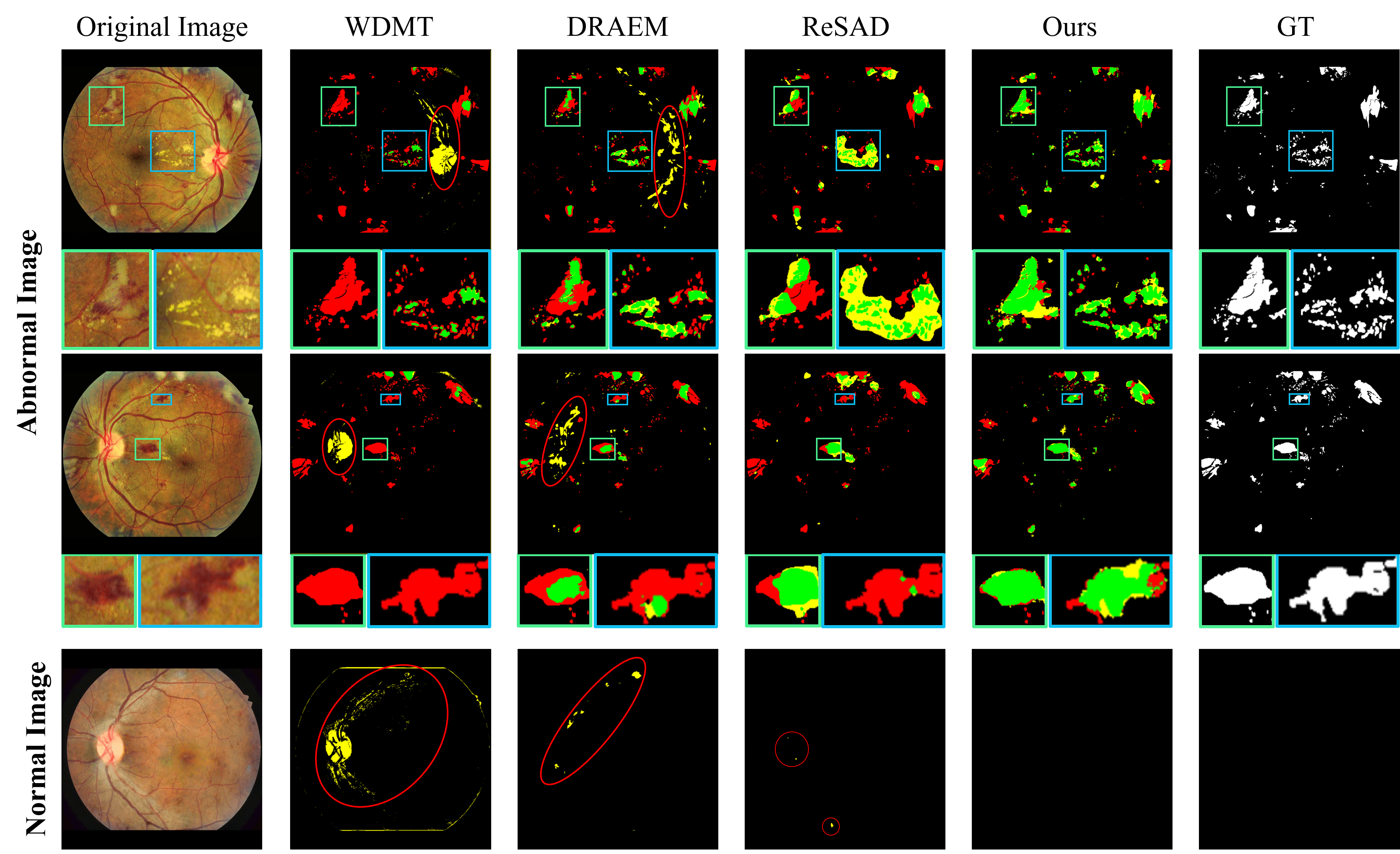}}
\caption{\textbf{Lesion Localization Results}: The green area in the figure represents true positives, the yellow area represents false positives, and the red area represents missing detection. Red circles denote the false positive in the normal structure of fundus}
\label{fig:Vis}
\end{figure*}

Tab. \ref {tab:result_pixel} quantitatively compares our model with recent approaches on the pixel-level anomaly localization. Compared to the best state-of-the-art methods, our approach shows improvements of 2.6\% in AUROC, 4\% in ACC, and a significant increase of 17.1\% in AUPR.  This suggests that our method achieves precise anomaly localization results for  retinal lesions (which is supported by  Fig. \ref {fig:Vis}). In summary, our approach has achieved the best performance in the pixel-level anomaly localization task for retinal lesions.

\begin{center}
\begin{table}[th]
\centering
\small 
\begin{tabular}{c|c|c|c|c|c|c|c|c}
\hline
\multicolumn{5}{c|}{ Ablation Method} & \multicolumn{4}{c}{Metric } \\ 
\hline
Loc.Net.  & \multicolumn{2}{c|}{ Rec.Net.}    & \multicolumn{2}{c|}{ Concatenate Type }  &\multicolumn{3}{c|}{ IDRiD } & EyeQ(0vsall) \\ 
\cline{2-9}
&Random & Train & Image & Feature&   AUROC & ACC & AUPR & AUROC \\
\hline
& &\checkmark & & &  0.801 & 0.719 & 0.083 & 0.485 \\
\checkmark  & &  & &  & 0.891 & 0.815  & 0.331 & 0.656 \\
\checkmark  & \checkmark & & & \checkmark   &  0.883 &0.817 & 0.267 & 0.682 \\
\checkmark  &  &\checkmark &\checkmark & & 0.819  & 0.742 & 0.149 & 0.581 \\
\textbf{\checkmark}   & & \textbf{\checkmark} & & \textbf{\checkmark} & \textbf{0.931}  & \textbf{0.859} & \textbf{0.427} &  \textbf{0.722} \\
\hline
\end{tabular}
\caption{ Ablation on Architecture and Concatenate Type (IDRiD for pixel-level and EyeQ for image-level): "Loc.Net." indicates the use of a localization network, "Rec.Net." indicates the use of a reconstruction network ("Random" for random initialization and "Train" for a trained reconstruction network on the proxy task). "Image" concatenates reconstructed images (similar to DRAEM), while "Feature" concatenates reconstructive features (like our proposed method).}
\vspace{-0.4cm}
\label{tab:ablation_Arch}
\end{table}
\end{center}

\vspace{-1cm}

\begin{center}
\begin{table}[th]
\centering
\begin{tabular}{c|c|c|c|c|c|c}
\hline
\multicolumn{3}{c|}{ Ablation Method} & \multicolumn{4}{c}{Metric } \\ 
\hline
Source Image & Self-Mix & Mask $\bm{P}$ & \multicolumn{3}{c|}{IDRiD } & EyeQ (0vsall)\\
\cline{4-7}
& & & AUROC & ACC & AUPR & AUROC \\
\hline
Texture & \checkmark &  \checkmark & 0.849 & 0.747 & 0.181 & 0.666 \\
 Fundus &\checkmark &  &0.781  & 0.720 & 0.043 & 0.560\\
 Fundus & & \checkmark &  0.890 & 0.819  & 0.274 & 0.682 \\
Fundus & \textbf{\checkmark} & \textbf{\checkmark} & \textbf{0.931}  & \textbf{0.859} & \textbf{0.427} & \textbf{0.722} \\
\hline
\end{tabular}
\caption{Ablation  on Anomaly Generation (IDRiD for pixel-level and EyeQ for image-level).}
\vspace{-1cm}
\label{AnoGen}
\end{table}
\end{center}

\vspace{-0.2cm}
\subsection{Ablation Study}
\label{ablation}
\textbf{Influences of network architectures.} Table \ref{tab:ablation_Arch} presents the results of ablation experiments on architecture and concatenate type. 
Without the localization network (first line), utilizing only the autoencoder in the first training stage and calculating reconstruction error as the anomaly score leads to a significant performance drop. This underscores the importance of introducing the localization network for synthetic anomaly localization. In the ablation experiment of the reconstruction network (second and third lines), the absence of the reconstruction network or using randomly initialized network leads to a notable performance drop, emphasizing the significance of the reconstruction network. Additionally, concatenating restored images (fourth line) results in a significant performance degradation, indicating that image concatenation may not be suitable for fundus anomaly detection. Comparing the results in the second line with the third and fourth lines, we observe that concatenating random features or restored images leads to a reduction in model performance, indicating the presence of misleading information  will misguide the model and decrease its overall performance.

\noindent\textbf{Influences of anomaly generation methods.} Table \ref{AnoGen} presents the results of ablation experiments on the anomaly generator. The "Source Image"  denotes the origin of the source image, where "Texture" represents the external texture dataset (DTD \cite{cimpoi2014describing}) used by DRAEM, and "Fundus" represents the normal retinal images used for training. The "Self-Mix" indicates whether the Self-Mix module is applied to combine source and target information, and the "Mask $\bm{P}$" indicates whether the Perlin mask $\bm{P}$ is used for anomaly generation.

The experimental results show that using an external texture dataset (DTD \cite{cimpoi2014describing}) instead of consistent normal fundus images leads to a decline in performance. Additionally, the absence of the Self-Mix module results in a noticeable performance decrease, and excluding the Perlin mask $\bm{P}$ leads to a significant drop in performance. These findings underscore the importance of our proposed approach, which leverages consistent source images, diverse masks, and an appropriate fusion method can greatly improve performance.

\vspace{-0.3cm}
\subsection{Qualitative Results}
\label{Qualitative}
As Shown in Fig. \ref{fig:Vis}. It can be seen that the existing reconstruction-based method WDMT and synthetic anomalies based method DRAEM would detect the normal structure of the fundus (like vessel, optic cup and disc) to anomalies, as indicated by the red circles in the figure. Besides, the representation-based method ReSAD  can not provide precise localization results, leading to more false positive (yellow) regions around the lesions. Compared to our method, the three baseline methods show more missing (red) areas and more false positive (yellow) areas around the normal structure or lesion in the fundus images. Moreover, all baseline methods will miss some subtle lesions which is hard to detect, while our method can detect various shapes and types of lesions better and provide relatively fine-grained localization results. 
Additionally, our method exhibits strong generalization capability on more types of lesions, as shown in the Supplementary. 

\vspace{-0.3cm}
\section{Conclusion}

\vspace{-0.3cm}
In this paper, we introduced the ReSynthDetect network, a novel approach for unsupervised anomaly detection in fundus images. Our method incorporated a novel anomaly generator that produces consistent synthetic anomalies.  Besides, we introduced a reconstruction network to extract reconstructive features, which were then fused with the localization network for synthetic anomaly localization. Our approach outperformed baseline methods on the IDRiD and EyeQ datasets, demonstrating its effectiveness in retina anomaly detection tasks.

\bibliography{egbib}

\begin{thebibliography}{32}
\providecommand{\natexlab}[1]{#1}
\providecommand{\url}[1]{\texttt{#1}}
\expandafter\ifx\csname urlstyle\endcsname\relax
  \providecommand{\doi}[1]{doi: #1}\else
  \providecommand{\doi}{doi: \begingroup \urlstyle{rm}\Url}\fi

\bibitem[Azhar and Khodra(2020)]{Azhar2020}
Annisa~Nurul Azhar and Masayu~Leylia Khodra.
\newblock {Fine-tuning Pretrained Multilingual BERT Model for Indonesian
  Aspect-based Sentiment Analysis}.
\newblock \emph{2020 7th International Conference on Advanced Informatics:
  Concepts, Theory and Applications, ICAICTA 2020}, pages 2980--2988, 2020.
\newblock \doi{10.1109/ICAICTA49861.2020.9428882}.

\bibitem[Baur et~al.(2021)Baur, Denner, Wiestler, Navab, and
  Albarqouni]{baur2021autoencoders}
Christoph Baur, Stefan Denner, Benedikt Wiestler, Nassir Navab, and Shadi
  Albarqouni.
\newblock Autoencoders for unsupervised anomaly segmentation in brain mr
  images: a comparative study.
\newblock \emph{Medical Image Analysis}, 69:\penalty0 101952, 2021.

\bibitem[Cimpoi et~al.(2014)Cimpoi, Maji, Kokkinos, Mohamed, and
  Vedaldi]{cimpoi2014describing}
Mircea Cimpoi, Subhransu Maji, Iasonas Kokkinos, Sammy Mohamed, and Andrea
  Vedaldi.
\newblock Describing textures in the wild.
\newblock In \emph{Proceedings of the IEEE conference on computer vision and
  pattern recognition}, pages 3606--3613, 2014.

\bibitem[Defard et~al.(2021)Defard, Setkov, Loesch, and
  Audigier]{defard2021padim}
Thomas Defard, Aleksandr Setkov, Angelique Loesch, and Romaric Audigier.
\newblock Padim: a patch distribution modeling framework for anomaly detection
  and localization.
\newblock In \emph{International Conference on Pattern Recognition}, pages
  475--489. Springer, 2021.

\bibitem[Emma~Dugas(2015)]{diabetic-retinopathy-detection}
Will~Cukierski Emma~Dugas, Jared~Jorge.
\newblock Diabetic retinopathy detection, 2015.
\newblock URL
  \url{https://kaggle.com/competitions/diabetic-retinopathy-detection}.

\bibitem[Felzenszwalb and Huttenlocher(2012)]{felzenszwalb2012distance}
Pedro~F Felzenszwalb and Daniel~P Huttenlocher.
\newblock Distance transforms of sampled functions.
\newblock \emph{Theory of computing}, 8\penalty0 (1):\penalty0 415--428, 2012.

\bibitem[Fu et~al.(2019)Fu, Wang, Shen, Cui, Xu, Liu, and
  Shao]{fu2019evaluation}
Huazhu Fu, Boyang Wang, Jianbing Shen, Shanshan Cui, Yanwu Xu, Jiang Liu, and
  Ling Shao.
\newblock Evaluation of retinal image quality assessment networks in different
  color-spaces.
\newblock In \emph{Medical Image Computing and Computer Assisted
  Intervention--MICCAI 2019: 22nd International Conference, Shenzhen, China,
  October 13--17, 2019, Proceedings, Part I 22}, pages 48--56. Springer, 2019.

\bibitem[Gong et~al.(2019)Gong, Liu, Le, Saha, Mansour, Venkatesh, and
  Hengel]{gong2019memorizing}
Dong Gong, Lingqiao Liu, Vuong Le, Budhaditya Saha, Moussa~Reda Mansour, Svetha
  Venkatesh, and Anton van~den Hengel.
\newblock Memorizing normality to detect anomaly: Memory-augmented deep
  autoencoder for unsupervised anomaly detection.
\newblock In \emph{Proceedings of the IEEE/CVF International Conference on
  Computer Vision}, pages 1705--1714, 2019.

\bibitem[Huang et~al.(2022)Huang, Huang, Luo, and Tang]{huang2022lesion2void}
Yijin Huang, Weikai Huang, Wenhao Luo, and Xiaoying Tang.
\newblock Lesion2void: unsupervised anomaly detection in fundus images.
\newblock In \emph{2022 IEEE 19th International Symposium on Biomedical Imaging
  (ISBI)}, pages 1--5. IEEE, 2022.

\bibitem[Huang et~al.(2023)Huang, Wang, Ye, Niu, Tu, Yang, Du, Deng, Gu, and
  He]{huang2023revisiting}
Ziyan Huang, Haoyu Wang, Jin Ye, Jingqi Niu, Can Tu, Yuncheng Yang, Shiyi Du,
  Zhongying Deng, Lixu Gu, and Junjun He.
\newblock Revisiting nnu-net for iterative pseudo labeling and efficient
  sliding window inference.
\newblock In \emph{Fast and Low-Resource Semi-supervised Abdominal Organ
  Segmentation: MICCAI 2022 Challenge, FLARE 2022, Held in Conjunction with
  MICCAI 2022, Singapore, September 22, 2022, Proceedings}, pages 178--189.
  Springer, 2023.

\bibitem[Kersting and {\v{Z}}elezn{\`y}(2013)]{kersting2013machine}
Hendrik Blockeel~Kristian Kersting and Siegfried Nijssen~Filip
  {\v{Z}}elezn{\`y}.
\newblock Machine learning and knowledge discovery in databases.
\newblock 2013.

\bibitem[Li et~al.(2021)Li, Sohn, Yoon, and Pfister]{li2021cutpaste}
Chun-Liang Li, Kihyuk Sohn, Jinsung Yoon, and Tomas Pfister.
\newblock Cutpaste: Self-supervised learning for anomaly detection and
  localization.
\newblock In \emph{Proceedings of the IEEE/CVF Conference on Computer Vision
  and Pattern Recognition}, pages 9664--9674, 2021.

\bibitem[Lin et~al.(2020)Lin, Li, Huang, Cheng, Xia, Wang, Yuan, and
  Tang]{lin2020sustech}
Li~Lin, Meng Li, Yijin Huang, Pujin Cheng, Honghui Xia, Kai Wang, Jin Yuan, and
  Xiaoying Tang.
\newblock The sustech-sysu dataset for automated exudate detection and diabetic
  retinopathy grading.
\newblock \emph{Scientific Data}, 7\penalty0 (1):\penalty0 409, 2020.

\bibitem[Mao et~al.(2020)Mao, Xue, Wang, Zhang, Zheng, and
  Liu]{mao2020abnormality}
Yifan Mao, Fei-Fei Xue, Ruixuan Wang, Jianguo Zhang, Wei-Shi Zheng, and Hongmei
  Liu.
\newblock Abnormality detection in chest x-ray images using uncertainty
  prediction autoencoders.
\newblock In \emph{International Conference on Medical Image Computing and
  Computer-Assisted Intervention}, pages 529--538. Springer, 2020.

\bibitem[Niu et~al.(2023)Niu, Dong, Yu, Dang, and Ding]{Niu2023}
Jingqi Niu, Shiwen Dong, Qinji Yu, Kang Dang, and Xiaowei Ding.
\newblock {Region and Spatial Aware Anomaly Detection for Fundus Images}.
\newblock pages 2--6, 2023.
\newblock URL \url{http://arxiv.org/abs/2303.03817}.

\bibitem[Ouardini et~al.(2019)Ouardini, Yang, Unnikrishnan, Romain, Garcin,
  Zenati, Campbell, Chiang, Kalpathy-Cramer, Chandrasekhar,
  et~al.]{ouardini2019towards}
Khalil Ouardini, Huijuan Yang, Balagopal Unnikrishnan, Manon Romain, Camille
  Garcin, Houssam Zenati, J~Peter Campbell, Michael~F Chiang, Jayashree
  Kalpathy-Cramer, Vijay Chandrasekhar, et~al.
\newblock Towards practical unsupervised anomaly detection on retinal images.
\newblock In \emph{Domain Adaptation and Representation Transfer and Medical
  Image Learning with Less Labels and Imperfect Data}, pages 225--234.
  Springer, 2019.

\bibitem[Perlin(1985)]{perlin1985image}
Ken Perlin.
\newblock An image synthesizer.
\newblock \emph{ACM Siggraph Computer Graphics}, 19\penalty0 (3):\penalty0
  287--296, 1985.

\bibitem[Porwal et~al.(2020)Porwal, Pachade, Kokare, Deshmukh, Son, Bae, Liu,
  Wang, Liu, Gao, et~al.]{porwal2020idrid}
Prasanna Porwal, Samiksha Pachade, Manesh Kokare, Girish Deshmukh, Jaemin Son,
  Woong Bae, Lihong Liu, Jianzong Wang, Xinhui Liu, Liangxin Gao, et~al.
\newblock Idrid: Diabetic retinopathy--segmentation and grading challenge.
\newblock \emph{Medical image analysis}, 59:\penalty0 101561, 2020.

\bibitem[Ronneberger et~al.(2015)Ronneberger, Fischer, and
  Brox]{ronneberger2015u}
Olaf Ronneberger, Philipp Fischer, and Thomas Brox.
\newblock U-net: Convolutional networks for biomedical image segmentation.
\newblock In \emph{Medical Image Computing and Computer-Assisted
  Intervention--MICCAI 2015: 18th International Conference, Munich, Germany,
  October 5-9, 2015, Proceedings, Part III 18}, pages 234--241. Springer, 2015.

\bibitem[Roth et~al.(2022)Roth, Pemula, Zepeda, Sch{\"o}lkopf, Brox, and
  Gehler]{roth2022towards}
Karsten Roth, Latha Pemula, Joaquin Zepeda, Bernhard Sch{\"o}lkopf, Thomas
  Brox, and Peter Gehler.
\newblock Towards total recall in industrial anomaly detection.
\newblock In \emph{Proceedings of the IEEE/CVF Conference on Computer Vision
  and Pattern Recognition}, pages 14318--14328, 2022.

\bibitem[Saito and Rehmsmeier(2015)]{saito2015precision}
Takaya Saito and Marc Rehmsmeier.
\newblock The precision-recall plot is more informative than the roc plot when
  evaluating binary classifiers on imbalanced datasets.
\newblock \emph{PloS one}, 10\penalty0 (3):\penalty0 e0118432, 2015.

\bibitem[Salehi et~al.(2021)Salehi, Sadjadi, Baselizadeh, Rohban, and
  Rabiee]{salehi2021multiresolution}
Mohammadreza Salehi, Niousha Sadjadi, Soroosh Baselizadeh, Mohammad~H Rohban,
  and Hamid~R Rabiee.
\newblock Multiresolution knowledge distillation for anomaly detection.
\newblock In \emph{Proceedings of the IEEE/CVF conference on computer vision
  and pattern recognition}, pages 14902--14912, 2021.

\bibitem[Schlegl et~al.(2019)Schlegl, Seeb{\"o}ck, Waldstein, Langs, and
  Schmidt-Erfurth]{schlegl2019f}
Thomas Schlegl, Philipp Seeb{\"o}ck, Sebastian~M Waldstein, Georg Langs, and
  Ursula Schmidt-Erfurth.
\newblock f-anogan: Fast unsupervised anomaly detection with generative
  adversarial networks.
\newblock \emph{Medical image analysis}, 54:\penalty0 30--44, 2019.

\bibitem[Schl{\"u}ter et~al.(2022)Schl{\"u}ter, Tan, Hou, and
  Kainz]{schluter2022natural}
Hannah~M Schl{\"u}ter, Jeremy Tan, Benjamin Hou, and Bernhard Kainz.
\newblock Natural synthetic anomalies for self-supervised anomaly detection and
  localization.
\newblock In \emph{Computer Vision--ECCV 2022: 17th European Conference, Tel
  Aviv, Israel, October 23--27, 2022, Proceedings, Part XXXI}, pages 474--489.
  Springer, 2022.

\bibitem[Singh et~al.(2020)Singh, Wang, Gupta, Goli, Padmanabhan, and
  Guly{\'a}s]{singh20203d}
Satya~P Singh, Lipo Wang, Sukrit Gupta, Haveesh Goli, Parasuraman Padmanabhan,
  and Bal{\'a}zs Guly{\'a}s.
\newblock 3d deep learning on medical images: a review.
\newblock \emph{Sensors}, 20\penalty0 (18):\penalty0 5097, 2020.

\bibitem[Zavrtanik et~al.(2021)Zavrtanik, Kristan, and
  Sko{\v{c}}aj]{zavrtanik2021draem}
Vitjan Zavrtanik, Matej Kristan, and Danijel Sko{\v{c}}aj.
\newblock Draem-a discriminatively trained reconstruction embedding for surface
  anomaly detection.
\newblock In \emph{Proceedings of the IEEE/CVF International Conference on
  Computer Vision}, pages 8330--8339, 2021.

\bibitem[Zhang et~al.(2022)Zhang, Sun, Li, Liu, He, Liu, and
  Zheng]{zhang2022multi}
Wentian Zhang, Xu~Sun, Yuexiang Li, Haozhe Liu, Nanjun He, Feng Liu, and Yefeng
  Zheng.
\newblock A multi-task network with weight decay skip connection training for
  anomaly detection in retinal fundus images.
\newblock In \emph{International Conference on Medical Image Computing and
  Computer-Assisted Intervention}, pages 656--666. Springer, 2022.

\bibitem[Zhao et~al.(2021)Zhao, Li, He, Ma, Fang, Li, and
  Zheng]{zhao2021anomaly}
He~Zhao, Yuexiang Li, Nanjun He, Kai Ma, Leyuan Fang, Huiqi Li, and Yefeng
  Zheng.
\newblock Anomaly detection for medical images using self-supervised and
  translation-consistent features.
\newblock \emph{IEEE Transactions on Medical Imaging}, 40\penalty0
  (12):\penalty0 3641--3651, 2021.

\bibitem[Zhou et~al.(2020)Zhou, Xiao, Yang, Cheng, Liu, Luo, Gu, Liu, and
  Gao]{zhou2020encoding}
Kang Zhou, Yuting Xiao, Jianlong Yang, Jun Cheng, Wen Liu, Weixin Luo, Zaiwang
  Gu, Jiang Liu, and Shenghua Gao.
\newblock Encoding structure-texture relation with p-net for anomaly detection
  in retinal images.
\newblock In \emph{Computer Vision--ECCV 2020: 16th European Conference,
  Glasgow, UK, August 23--28, 2020, Proceedings, Part XX 16}, pages 360--377.
  Springer, 2020.

\bibitem[Zhou et~al.(2021)Zhou, Greenspan, Davatzikos, Duncan, Van~Ginneken,
  Madabhushi, Prince, Rueckert, and Summers]{zhou2021review}
S~Kevin Zhou, Hayit Greenspan, Christos Davatzikos, James~S Duncan, Bram
  Van~Ginneken, Anant Madabhushi, Jerry~L Prince, Daniel Rueckert, and Ronald~M
  Summers.
\newblock A review of deep learning in medical imaging: Imaging traits,
  technology trends, case studies with progress highlights, and future
  promises.
\newblock \emph{Proceedings of the IEEE}, 109\penalty0 (5):\penalty0 820--838,
  2021.

\bibitem[Zhu et~al.(2022)Zhu, Wang, Yin, Yang, Liao, and Li]{zhu2022selfmix}
Qikui Zhu, Yanqing Wang, Lei Yin, Jiancheng Yang, Fei Liao, and Shuo Li.
\newblock Selfmix: A self-adaptive data augmentation method for lesion
  segmentation.
\newblock In \emph{Medical Image Computing and Computer Assisted
  Intervention--MICCAI 2022: 25th International Conference, Singapore,
  September 18--22, 2022, Proceedings, Part IV}, pages 683--692. Springer,
  2022.

\bibitem[Zuiderveld(1994)]{zuiderveld1994contrast}
Karel Zuiderveld.
\newblock Contrast limited adaptive histogram equalization.
\newblock \emph{Graphics gems}, pages 474--485, 1994.

\end{thebibliography}
\end{document}